\pdfoutput=1

\documentclass[11pt]{article}

\usepackage[]{acl}

\usepackage{times}
\usepackage{latexsym}

\usepackage[T1]{fontenc}

\usepackage[utf8]{inputenc}

\usepackage{microtype}

\usepackage{booktabs}
\usepackage{adjustbox}
\usepackage{xspace}
\usepackage{multirow}
\usepackage{dsfont}
\usepackage{contour}
\usepackage{textcomp}  
\usepackage{scalerel}  
\usepackage{tabularx}

\usepackage{amsfonts}
\usepackage{refstyle}

\usepackage{amsmath}
\usepackage{soul}

\usepackage{colortbl}

\newtheorem{mydef}{Definition}

\newcommand{\CSadj}{commonsense\xspace}
\newcommand{\CSn}{common sense\xspace}

\newcommand{\QT}[1]{``{#1}''\xspace}
\newcommand{\CQ}{\textit{PaCo}\xspace}

\newcommand{\weakcq}{\textit{PInKS}\xspace}
\newcommand{\pabi}{\textit{PABI}\xspace}

\newcommand{\recall}{precision\xspace}
\newcommand{\recalls}{precisions\xspace}
\newcommand{\Recall}{Precision\xspace}

\usepackage{cleveref}
\crefformat{section}{\S#2#1#3}
\crefformat{subsection}{\S#2#1#3}
\crefformat{subsubsection}{\S#2#1#3}
\crefrangeformat{section}{\S\S#3#1#4 to~#5#2#6}
\crefmultiformat{section}{\S\S#2#1#3}{ and~#2#1#3}{, #2#1#3}{ and~#2#1#3}

\usepackage{graphicx}
\Crefformat{figure}{#2Fig.~#1#3}
\Crefmultiformat{figure}{Figs.~#2#1#3}{ and~#2#1#3}{, #2#1#3}{ and~#2#1#3}
\Crefformat{table}{#2Tab.~#1#3}
\Crefmultiformat{table}{Tabs.~#2#1#3}{ and~#2#1#3}{, #2#1#3}{ and~#2#1#3}
\Crefformat{appendix}{Appx.~\S#2#1#3}
\crefformat{algorithm}{Alg.~#2#1#3}

\newcommand{\fact}{statement\xspace}
\newcommand{\facts}{statements\xspace}

\newcommand{\fmac}{Macro-F1\xspace}

\definecolor{USCgold}{HTML}{F6C400}
\definecolor{USCred}{HTML}{990000}
\definecolor{blossomPink}{HTML}{FFD1EF}
\definecolor{blossomRed}{HTML}{EC5F90}

\definecolor{textgreen}{HTML}{26580F}
\definecolor{textred}{HTML}{7C0A02}

\newcommand{\blpinked}[1]{{\setlength{\fboxsep}{2pt}\hspace{-2pt}\colorbox{blossomPink}{#1}}}
\newcommand{\blreded}[1]{{\setlength{\fboxsep}{2pt}\hspace{-2pt}\colorbox{blossomRed}{#1}}}

\newcommand{\greened}[1]{{\color[HTML]{378805} #1}}
\newcommand{\reded}[1]{{\color[HTML]{E32227} #1}}

\title{\mbox{\contour{blossomRed}{\textcolor{blossomPink}{\weakcq}}: Preconditioned  Commonsense Inference with Weak Supervision
}}

\author{Ehsan Qasemi{$^{1}$}
    \and Piyush Khanna{$^{2}$}
    \and Qiang Ning{$^{3}$}
    \and Muhao Chen{$^{1}$}
    \\
   $^1$University of Southern California
    \;
    $^2$Delhi Technological University
    \;
    $^3$Amazon
    \\
    \texttt{\{qasemi,muhaoche\}@usc.edu}; \texttt{piyushkhanna\_bt2k17@dtu.ac.in};\\
    \texttt{qning@amazon.com}
    \\}

\date{}

\begin{document}
    \maketitle


\begin{abstract}
Reasoning with preconditions such as \QT{glass can be used for drinking water unless the glass is shattered} remains an open problem for language models.
The main challenge lies in the scarcity of preconditions data and model's lack of support for such reasoning.
We present \weakcq, \underline{P}reconditioned Commonsense \underline{In}ference with Wea\underline{K} \underline{S}upervision, an improved model for reasoning with preconditions through minimum supervision.
We show, both empirically and theoretically, that \weakcq improves the results on benchmarks focused on reasoning with the preconditions of commonsense knowledge~(up to $40\%$ \fmac scores).
We further investigate \weakcq through PAC-Bayesian informativeness analysis, \recall measures, and ablation study.\footnote{Code and data on \hyperlink{https://github.com/luka-group/PInKS}{https://github.com/luka-group/PInKS}}
\end{abstract}

    \section{Introduction}
    \label{sec:introduction}

\begin{figure}[ht]
    \centering
    \includegraphics[width=.8\columnwidth]{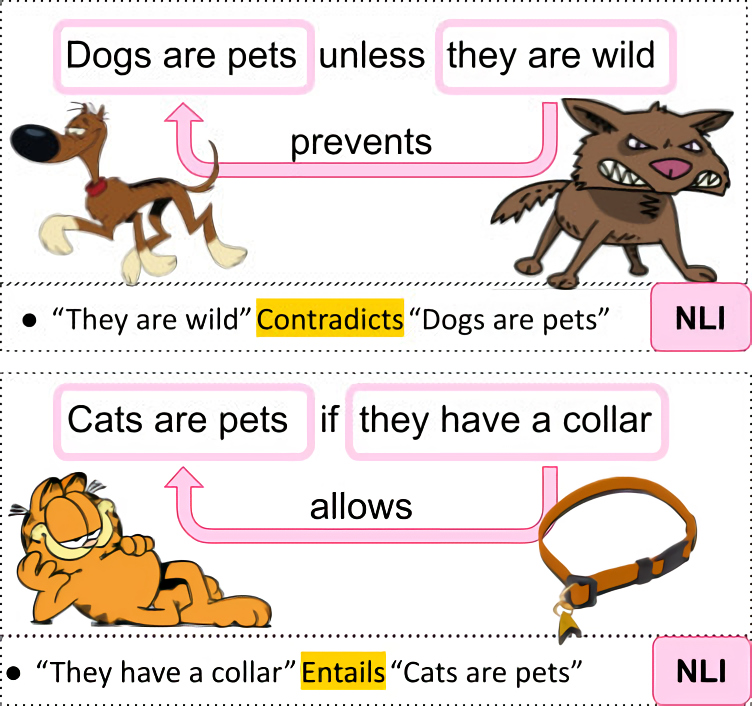}
    \caption{Examples on Preconditioned Inference and the NLI format they can be represented in. 
    }
    \label{fig:overview}
    \vspace{-1.5em}
\end{figure}

Inferring the effect of a situation or precondition on a subsequent action or state (illustrated in \Cref{fig:overview}) is an open part of \CSadj reasoning.
It requires an agent to possess and understand different dimensions of commonsense knowledge \cite{woodward2011psychological}, e.g.\ physical, causal, social, etc.
This ability can improve many knowledge-driven tasks such as question answering~\cite{wang2019superglue,talmor2018commonsenseqa}, machine reading comprehension~\cite{sakaguchi2020winogrande}, and narrative prediction~\cite{mostafazadeh-etal-2016-corpus}.
It also seeks to benefit a wide range of real-world intelligent applications such as legal document processing~\cite{hage2005law}, claim verification~\cite{nie2019combining}, and debate processing~\cite{widmoser-etal-2021-randomized}.

Multiple recent studies have taken the effort on reasoning with preconditions of \CSadj knowledge \cite{rudinger2020thinking,qasemi2021corequisite,mostafazadeh2020glucose,hwang2020comet}.
These studies show that preconditioned reasoning represents an unresolved challenge to state-of-the-art (SOTA) language model (LM) based reasoners.
Generally speaking, the problem of reasoning with preconditions has been formulated as variations of the natural language inference (NLI) task where, given a precondition/update, the model has to decide its effect on a \CSn statement or chain of statements.
For example, \CQ~\cite{qasemi2021corequisite} approaches the task from the causal (hard reasoning) perspective in term of \emph{enabling} and \emph{disabling} preconditions of \CSadj knowledge, and evaluate reasoners with crowdsourced \CSadj statements about the two polarities of preconditions of statements in ConceptNet~\cite{speer2016conceptnet}.
Similarly, $\delta-$NLI~\cite{rudinger2020thinking} formulates the problem
from soft assumptions' perspective, i.e., \emph{weakeners} and \emph{strengtheners}, and justifies whether the \emph{update} sentence \emph{weakens} or \emph{strengthens} the textual entailment in sentence pairs from sources such as SNLI~\cite{snli}.
Obviously, both tasks capture the same phenomena of reasoning with preconditions and the slight difference in format does not hinder their usefulness~\cite{gardner2019question}.
As both works conclude, SOTA models generally fall short of tackling these tasks.

We identify two reasons for such shortcomings of LMs on reasoning with preconditions: 1) relying on expensive direct supervision and 2) the need for improved LMs to reason with such knowledge.
First, current resources for preconditions of \CSn are manually annotated. 
Although this yields high-quality direct supervision, it is costly and not scalable.
Second, off-the-shelf LMs are trained on free-text corpora with no direct guidance on specific tasks.
Although such models can be further fine-tuned to achieve impressive performance on a wide range of tasks, they are far from perfect in reasoning on preconditions due to their complexity of need for deep \CSadj understanding and lack of large-scale training data.

\begin{figure*}[ht]
    \centering
    \includegraphics[width=0.9\textwidth]{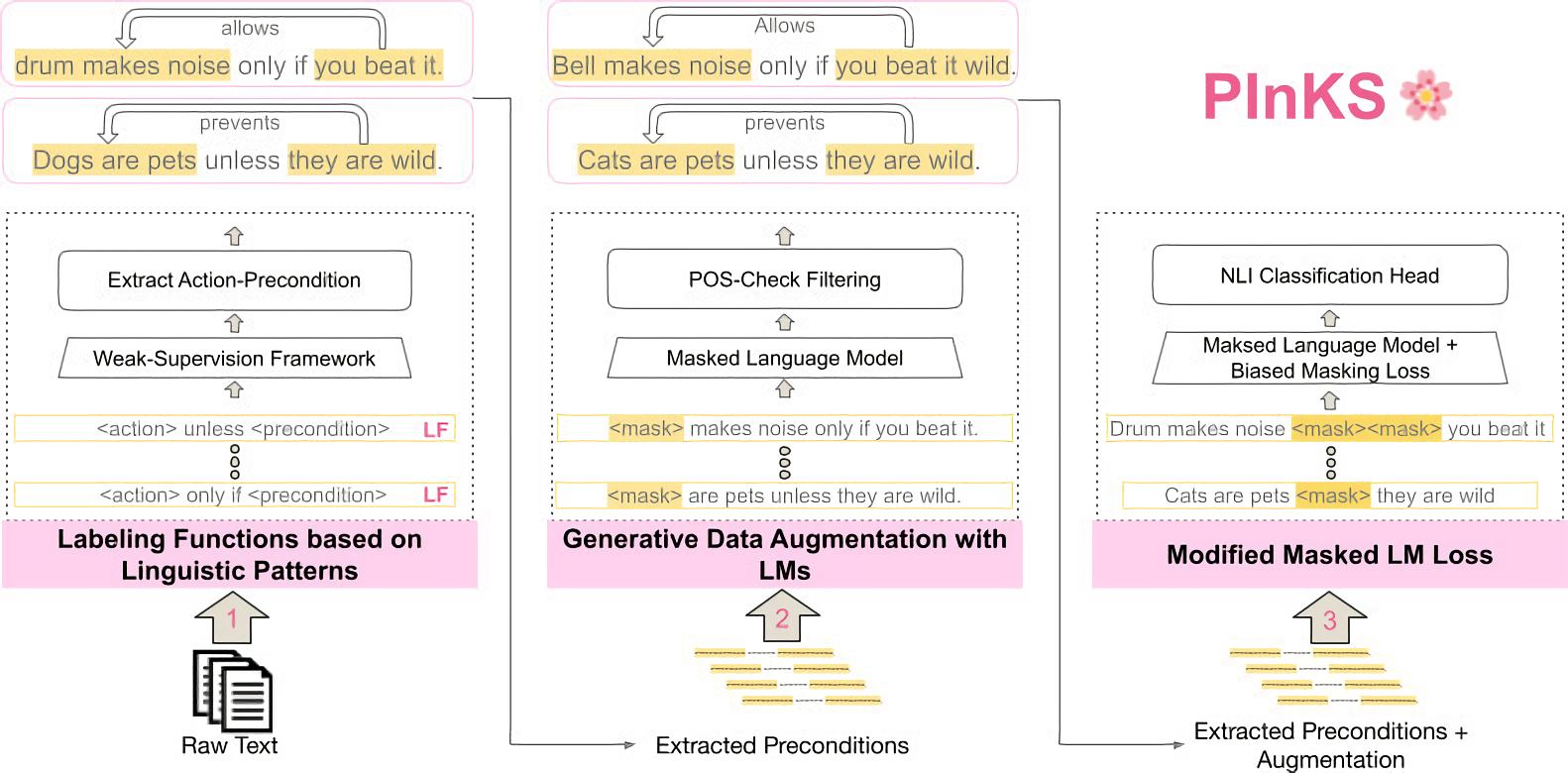}
    \caption{Overview of the three minimally supervised methods in \weakcq.
    }
    \label{fig:pinks}
    
    \vspace{-1.5em}
\end{figure*}

In this work, we present \weakcq (see \Cref{fig:pinks}), a minimally supervised approach for reasoning with the precondition of \CSadj knowledge in LMs.
The main contributions are 3 points.
\textbf{First}, to enhance training of the reasoning model~(\Cref{sec:weakcq}), we propose two strategies of retrieving rich amount of cheap supervision signals (\Cref{fig:overview}).
In the first strategy~(\Cref{subsec:labeling}), we use common linguistic patterns (e.g. \QT{[action] unless [precondition]}) to gather sentences describing preconditions and actions associated with them from massive free-text corpora (e.g.\ OMCS~\cite{omcs}).
The second strategy~(\Cref{subsec:augmentation}) then uses generative data augmentation methods on top of the extracted sentences to induce even more training instances.
As the \textbf{second} contribution~(\Cref{subsec:modified-mlm}), we improve LMs with more targeted preconditioned commonsense inference.
We modify the masked language model (MLM) learning objective to biased masking, which puts more emphasis on preconditions, hence improving the LMs capability to reason with preconditions.
Finally, for \textbf{third} contribution, we go beyond empirical analysis of \weakcq and investigate the performance and robustness through theoretical guarantees of PAC-Bayesian analysis~\cite{forseeing}.

Through extensive evaluation on five representative datasets~(ATOMIC2020~\cite{hwang2020comet}, WINOVENTI~\cite{do2021rotten}, ANION~\cite{jiang2021m}, \CQ~\cite{qasemi2021corequisite} and DNLI~\cite{rudinger2020thinking}), we show that \weakcq improves the performance of NLI models, up to $5\%$ \fmac without seeing any task-specific training data and up to $40\%$ \fmac after being incorporated into them (\Cref{subsec:target-tasks}).
In addition to the empirical results, using theoretical guarantees of informativeness measure in \pabi~\cite{forseeing}, we show that the minimally supervised data of \weakcq is as informative as fully supervised datasets~(\Cref{subsec:info-eval}).
Finally, to investigate the robustness of \weakcq and effect of each component, we focus on the weak supervision part (\Cref{sec:datastats}). 
We perform ablation study of \weakcq w.r.t. the linguistic patterns themselves, the recall value associated with linguistic patterns, and finally contribution of each section to overall quality and the final performance. 

    \section{Problem Definition}
    \label{sec:background}
    
%

Common sense \facts describe well-known information about concepts, and, as such, they are acceptable by people without need for debate~\cite{sap2019atomic,davis2015commonsense}.
The preconditions of \CSn knowledge are eventualities that affect happening of a \CSn \fact~\cite{hobbs2005toward}.
These preconditions can either \textit{allow} or \textit{prevent} the \CSn \fact in different degrees~\cite{rudinger2020thinking,qasemi2021corequisite}. 
For example, \citet{qasemi2021corequisite} model the preconditions as \emph{enabling} and \emph{disabling} (hard preconditions), whereas \citet{rudinger2020thinking} model them as \emph{strengthening} and \emph{weakening}(soft preconditions).
Beyond the definition of preconditions, the task of inference with preconditions is also defined differently among the literature. 
Some task definitions have strict constraints on the format of statement, e.g. two sentence format ~\cite{rudinger2020thinking} or being human-related \cite{sap2019atomic}, whereas others do not~\cite{do2021rotten,qasemi2021corequisite}. 

To unify the definitions in available literature, we define the preconditioned inference task as below:
\begin{mydef}
    \textbf{Preconditioned Inference:} \label{def:precondition}
    given a \CSn \fact and an update sentence that serves as precondition, is the \fact still allowed or prevented?
\end{mydef}
This definition is consistent with definitions in the literature (for more details see \cref{sec:other_prec_inference}).
First, similar to the definition by \citet{rudinger2020thinking}, the update can have different levels of effect on the statement, from causal connection (hard) to material implication (soft).
Second, similar to the one \citet{qasemi2021corequisite}, the statement can have any format. 


    \section{Preconditioned Inference with Minimal Supervision}
    \label{sec:weakcq}
In \weakcq, to overcome the challenges associated with inference with preconditions, we propose two sources of weak supervision to enhance the training of a reasoner: linguistic patterns to gather rich (but allowably noisy) preconditions (\Cref{subsec:labeling}), and generative augmentation of the preconditions data (\Cref{subsec:augmentation}).
The main hypothesis in using weak-supervision methods is that pretraining models on large amount of weak-supervised labeled data could improve model's performance on similar downstream tasks~\cite{ratner2017snorkel}.
In weak supervision terminology for heuristics, the experts design a set of heuristic labeling functions (LFs) that serves as the generators of the noisy label~\cite{ratner2017snorkel}.
These labeling functions can produce overlapping or conflicting labels for a single instance of data that will need to be resolved either with simple methods such as ensemble inference or more sophisticated probabilistic methods such as data programming~\cite{ratner2016data}, or generative~\cite{bach2017learning}.
Here, the expert still needs to design the heuristics to query the knowledge and convert the results to appropriate labels for the task.
In addition, we propose the modified language modeling objective that uses biased masking to improve the precondition-reasoning capabilities of LMs (\Cref{subsec:modified-mlm}).

\subsection{Weak Supervision with Linguistic Patterns}
\label{subsec:labeling}
We curate a large-scale automatically labeled dataset for, both type of, preconditions of commonsense statements by defining a set of linguistic patterns and searching through raw corpora.
Finally, we have a post-processing filtering step to ensure the quality of the extracted preconditions.

\paragraph{Raw Text Corpora:}
In our experiments, we acquire weak supervision from two corpora: Open Mind Common Sense (OMCS)~\cite{singh2002open} and ASCENT~\cite{nguyen2021advanced}.
OMCS is a large commonsense statement corpus that contains over 1M sentences from over 15,000 contributors.
ASCENT has consolidated over 8.9M commonsense statements from the Web. 

First, we use sentence tokenization in NLTK~\cite{nltk} to separate individual sentences in the raw text. 
Each sentence is then considered as an individual statement to be fed into the labeling functions.
We further filter out the data instances based on the conjunctions used in the common sense statements after processing the labeling functions~(discussed in Post-Processing paragraph).

\begin{table*}[ht]
    \centering
    \small
        \begin{tabularx}{\textwidth}{X|l|X|l}
            \midrule
            \textbf{Text} 
                & \textbf{Label} 
                & \textbf{Action} 
                & \textbf{Precondition} 
            \\ \toprule
            A drum makes noise \greened{only if} you beat it.
                & \greened{Allow}  
                & A drum makes noise
                & you beat it.          
            \\ 
            Your feet might come into contact with something \greened{if} it is on the floor. 
                & \greened{Allow}  
                & Your feet might come into contact with something 
                & it is on the floor.   
            \\ 
            Pears will rot \reded{if not} refrigerated       
                & \reded{Prevent}
                & Pears will rot
                & refrigerated
            \\ 
            Swimming pools have cold water in the winter \reded{unless} they are heated.
                & \reded{Prevent}
                & Swimming pools have cold water in the winter
                & they are heated. 
            \\ \bottomrule    
        \end{tabularx}
    \caption{Examples from the collected dataset through linguistic patterns in \Cref{subsec:labeling}.
    }
    \label{tab:examples}
    \vspace{-1.5em}
\end{table*}

\paragraph{Labeling Functions (LF):}
We design the LFs required for weak-supervision with a focus on the presence of a linguistic pattern in the sentences based on a conjunction (see \Cref{tab:examples} for examples).
In this setup, each LF labels the training data as \emph{Allowing}, \emph{Preventing} or \emph{Abstaining} (no label assigned) depending on the linguistic pattern it is based on. 
For example, as shown in \Cref{tab:examples} the presence of conjunctions \textit{only if} and \textit{if}, with a specific pattern, suggests that the precondition \emph{Allows} the action. 
Similarly, the presence of the conjunction \textit{unless} indicates a \emph{Preventing} precondition. 
We designed 20 such LFs based on individual conjunctions through manual inspection of the collected data in several iterations, for which details are described in~\cref{subsec:appendx-patterns}. 

\paragraph{Extracting Action-Precondition Pairs}
\label{subsec:extracting-action-prec}
Once the sentence have an assigned label, we extract the \textit{action-precondition} pairs using the same linguistic patterns.
This extraction can be achieved by leveraging the fact that a conjunction divides a sentence into \textit{action} and \textit{precondition} in the following pattern  \QT{\textit{precondition} \textit{conjunction} \textit{action}}, as shown in \Cref{tab:examples}.

However, there could be sentences that contain multiple conjunctions. 
For instance, the sentence \QT{Trees continue to grow for all their lives except in winter if they are not evergreen.} includes two conjunctions \QT{except} and \QT{if}.
Such co-occurring conjunctions in a sentence leads to ambiguity in the extraction process.
To overcome this challenge, we further make selection on the patterns by measuring their \recalls\footnote{The amounts of labeled instances (\textit{non-abstaining}) for each labeling function are relevant}.
To do so, we sample $20$ random sentences from each conjunction ($400$ total) and label them manually on whether they are relevant to our task or not by two expert annotators. 
If a sentence is relevant to the task, it is labeled as $1$; otherwise, $0$.
We then average the scores of two annotators for each pattern/conjunction to get its \recall score. 
This \recall score serves as an indicator of the quality of preconditions extracted by the pattern/conjunction in the context of our problem statement.
Hence, priority is given to a conjunction with a higher \recall in case of ambiguity. 
Further, we also set a minimum \recall threshold (=$0.7$) to filter out the conjunctions having a low \recall score (8 LFs), indicating low relevance to the task of reasoning with preconditions (see \Cref{subsec:appendx-patterns} for list of \recall values).


\paragraph{Post-Processing}
On manual inspection of sentences matched by the patterns, we observed a few instances from random samples that were not relevant to the context of commonsense reasoning tasks, for example: \textit{How do I know if he is sick?} or, \textit{Pianos are large but entertaining}.
We accordingly filter out sentences that are likely to be irrelevant instances. Specifically, those include 1) questions which are identified based on presence of question mark and interrogative words (List of interrogative words in \Cref{subsec:interrogative-words}), or 2) do not have a verb in their precondition. 
Through this process we end up with a total of 113,395 labeled action-precondition pairs with 102,474 \emph{Allow} and 10,921 \emph{Prevent} assertions.


\subsection{Generative Data Augmentation}
\label{subsec:augmentation}

To further augment and diversify training data, we leverage another technique of retrieving weak supervision signals by probing LMs for generative data augmentation.
To do so, we mask the nouns and adjectives (pivot-words) from the text and let the generative language model fill in the masks with appropriate alternatives.

After masking the pivot-word and filling in the mask using the LM, we filter out the augmentations that change the POS tag of the pivot-word and then keep the top 3 predictions for each mask.
In addition, to keep the diversity of the augmented data, we do not use more than 20 augmented sentences for each original statement (picked randomly).
For example, in the statement \QT{Dogs are pets unless they are wild}, the pivot-words are \QT{dogs}, \QT{pets} and \QT{wild}.
Upon masking \QT{dogs}, using RoBERTa (large) language model, we get valid augmentations such as \QT{\underline{Cats} are pets unless they are wild}.
Using this generative data augmentation, we end up with $7M$ labeled action-precondition pair with $11\%$ \emph{prevent} preconditions.

\subsection{Precondition-Aware Biased Masking}
\label{subsec:modified-mlm}
To increase the LM's attention on preconditions, we used biased masking on conjunctions as the closest proxies to preconditions' reasoning.
Based on this observation, we devised a biased masked language modeling loss that solely focuses on masking conjunctions in the sentences instead of random tokens.
Similar to \citet{dai2019transformer}, we mask the whole conjunction word in the sentence and ask the LM to fulfill the mask.
The goal here is to start from a pretrained language model and, through this additional fine-tuning step, improve its ability to reason with preconditions.
To use such fine-tuned LM in a NLI module, we further fine-tune the \QT{LM+classification head} on subset of MNLI~\cite{mnli} dataset.
For full list of conjunctions and implementation details check \Cref{subsec:appendx-ModMLM}.

    \section{Experiments}
    \label{sec:experiments}
    

This section first showcases improvements of \weakcq on five representative tasks for preconditioned inference (\Cref{subsec:target-tasks}).
We then theoretically justify the improvements by measuring the informativeness of weak supervision by \weakcq using \pabi~\cite{forseeing} score and then experiment on the effect of \recall (discussed in \cref{subsec:extracting-action-prec}) on \weakcq using \pabi score (\Cref{subsec:info-eval}).
Additional analysis on various training strategies of \weakcq is also provided in \Cref{subsec:curic-vs-multitask}.

\subsection{Main Results}
\label{subsec:target-tasks}
Comparing the capability for models to reason with preconditions across different tasks requires canonicalizing the inputs and outputs in such tasks be in the same format. 
We used natural language inference (NLI) as such a canonical format. 
\CQ~\cite{qasemi2021corequisite} and $\delta$-NLI~\cite{rudinger2020thinking} are already formulated as NLI and others can be converted easily using the groundwork laid by \citet{qasemi2021corequisite}.
In NLI, given a sentence pair with a \emph{hypothesis} and a \emph{premise}, one predicts whether the hypothesis is true (entailment), false (contradiction), or undetermined (neutral) given the premise~\cite{mnli}.
Each task is preserved with equivalence before and after any format conversion at here, hence conversion does not seek to affect the task performance, inasmuch as it is discussed by \citet{gardner2019question}.
More details on this conversion process are in \Cref{appendix:target-datasets}, and examples from the original target datasets are given in \Cref{tab:target-to-NLI}.

\paragraph{Setup}
To implement and execute labeling functions, and resolve labeling conflict, we use Snorkel~\cite{ratner2017snorkel}, one of the SOTA frameworks for algorithmic labeling on raw data that provides ease-of-use APIs.\footnote{Other alternatives such as skweak~\cite{lison2021skweak} can also be used for this process.}
For more details on Snorkel and its setup details, please see Appendix~\ref{subsec:appendx-snorkel}.

For each target task, we start from a pretrained NLI model (RoBERTa-Large-MNLI~\cite{liu2019roberta}), fine-tune it according to \weakcq (as discussed in \Cref{sec:weakcq}) and evaluate its performance on the test portion of the target dataset in two setups: zero-shot transfer learning without using the training data for the target task (labeled as \emph{\weakcq} column) and fine-tuned on the training portion of the target task (labeled as \emph{Orig.+\weakcq}).
To facilitate comparison, we also provide the results for fully fine-tuning on the training portion of the target task and evaluating on its testing portion (labeled as \emph{Orig.} column; \weakcq is not used here).
To create the test set, if the original data does not provide a split (e.g. \textit{ATOMIC} and \textit{Winoventi}), following \citet{qasemi2021corequisite}, we use unified random sampling with the $[0.45, 0.15, 0.40]$ ratio for train/dev/test.
The experiments are conducted on a commodity workstation with an Intel Xeon Gold 5217 CPU and an NVIDIA RTX 8000 GPU.
For all the tasks, we used the pretrained model from \emph{huggingface}~\cite{huggingface}, and utilized PyTorch Lightning~\cite{Falcon_PyTorch_Lightning_2019} library to manage the fine-tuning process.
We evaluate each performance by aggregating the \emph{\fmac} score (implemented in \citet{sklearn}) on the ground-truth labels and report the results on the unseen test split of the data.

\paragraph{Discussion}
\begin{table}[ht]
    \small
    \centering
    \begin{tabular}{l|l|ll}
    \midrule
     \textbf{Target Task}
         & \textbf{Orig.}
         & \textbf{\weakcq}
         & \textbf{Orig+\weakcq}
     \\
     \toprule
     \textit{$\delta$-NLI} & 83.4 & 60.3 & \blpinked{\textbf{84.1}} \\
     \CQ           & 77.1 & 69.5 & \blpinked{\textbf{79.4}}\\
     \textit{ANION}        & 81.1 & 52.9 & \blpinked{\textbf{81.2}}\\
     \textit{ATOMIC}       & 43.2 & \blpinked{\textbf{48.0}} & \blpinked{\textbf{88.6}}\\
     \textit{Winoventi}    & 51.1 & \blpinked{\textbf{52.4}} & \blpinked{\textbf{51.3}}\\
     \bottomrule
    \end{tabular}
    \caption{\fmac (\%) results of \weakcq on the target datasets: no \weakcq (\emph{Orig.}), with \weakcq in zero-shot transfer learning setup (\emph{\weakcq}) and \weakcq in addition to original task's data (\emph{Orig.+\weakcq}). \blpinked{\textbf{Bold}} values are cases where \weakcq is improving supervised results. 
    }
    \label{tab:target-eval}
    \vspace{-1em}
\end{table}
Table~\ref{tab:target-eval} presents the evaluation results of this section.
As illustrated, on \textit{ATOMIC}~\cite{hwang2020comet} and \textit{Winoventi}~\cite{do2021rotten}, \weakcq exceeds the supervised results even without seeing any examples from the target data (zero-shot transfer learning setup).
On \textit{$\delta$-NLI}~\cite{rudinger2020thinking}, \textit{ANION}~\cite{jiang2021m} and \textit{ATOMIC}~\cite{hwang2020comet}, combination of \weakcq and train subset of target task (\weakcq in low-resource setup) outperforms the target task results. 
This shows \weakcq can also utilize additional data from target task to achieve better performance consistently across different aspects of preconditioned inference.

\subsection{Informativeness Evaluation}
\label{subsec:info-eval}
\citet{forseeing} proposed a unified PAC-Bayesian motivated informativeness measure, namely \pabi, that correlates with the improvement provided by the incidental signals to indicate their effectiveness on a target task.
The incidental signal can include an inductive signal, e.g.\ partial/noisy labeled data, or a transductive signal, e.g.\ cross-domain signal in transfer learning.

In this experiment, we go beyond the empirical results and use the \pabi measure to explain how improvements from \weakcq are theoretically justified.
Here, we use the \pabi score for cross-domain signal assuming the weak supervised data portion of \weakcq (\Cref{subsec:labeling} and \Cref{subsec:augmentation}) as a indirect signal for a given target task.
We use \pabi measurements from two perspective. First, we examine how useful is the weak supervised data portion of \weakcq for target tasks in comparison with fully-supervised data. And second, we examine how the \recall of the linguistic patterns (discussed in \Cref{subsec:extracting-action-prec}) affects this usefulness.

\paragraph{Setup}
We carry over the setup on models and tasks from~\Cref{subsec:target-tasks}.
For details on the \pabi itself and the measurement details associated with it, please see~\Cref{sec:details-on-pabi}.
For the aforementioned first perspective, we only consider \CQ and $\delta$-NLI as target tasks, as they are the two main learning resources specifically focused on preconditioned inference (as defined in Section 2), which is not the case for others.
We measure the \pabi of the weak supervised data portion of \weakcq on the two target tasks, and compare it with the \pabi of the fully-supervised data from ~\Cref{subsec:target-tasks}.
For the second perspective, we only focus on \weakcq and consider \CQ as target task. 
We create different versions of the weak supervised data portion of \weakcq with different levels of \recall threshold (e.g. $0.0$, $0.5$) and compare their informativeness on \CQ.
To limit the computation time, we only use $100K$ samples from the weak supervised data portion of \weakcq in each threshold value, which is especially important in lower thresholds due to huge size of extracted patterns with low \recall threshold.

\paragraph{Informativeness in Comparison with Direct Supervision:}
\Cref{tab:pabi-eval} summarizes the \pabi informativeness measure in comparison with other datasets with respect to \CQ~\cite{qasemi2021corequisite} and $\delta$-NLI~\cite{rudinger2020thinking}.
To facilitate the comparison of \pabi scores in \Cref{tab:pabi-eval}, we have also reported the minimum achievable (\QT{zero rate} classifier) and maximum achievable \pabi scores.
To clarify, to compute the maximum achievable \pabi score, we consider the training subset of the target task as an indirect signal for the test subset. 
Here, we assume that the training subset is in practice the most informative indirect signal available for the test subset of any task.
For the minimum achievable \pabi score, we considered the error rate of the \QT{zero rate} classifier (always classifies to the largest class) for computations of \pabi.

Our results show that although, \weakcq is the top informative incidental signal in $\delta$-NLI target task and second best in \CQ (less than 0.001 point of difference with the best signal). This \pabi numbers are even more significant considering that \weakcq is the only weak-supervision data which is automatically acquired, while others are acquired through sometimes multiple rounds of human annotations and verification.

\begin{table}[t]
    \small
    \centering
    \setlength\tabcolsep{1pt}
    \begin{tabularx}{\linewidth}{l|cc|X}
         \hline
         & \multicolumn{2}{c|}{\textbf{\pabi on}} &
        \\
            
            \textbf{Indir. Task}
            & \textbf{\CQ}
            & \textbf{$\delta$-NLI}
            & \textbf{Explanation}
        \\
        \toprule
        \weakcq      & 52.2 & \blpinked{\textit{66.7}} & - Best on $\delta$-NLI \\ \hline
        $\delta$-NLI & \blpinked{\textit{52.3}} & \blreded{\textbf{85.5}} & - Max achievable on $\delta$-NLI \\ &&& - Best on \CQ  \\
        \CQ           & \blreded{\textbf{52.3}} & 31.3 & - Max achievable on \\&&&\CQ \\
        ANION        & 34.1 & 13.9\\
        ATOMIC       & 20.9 & 17.4 \\
        Winoventi    & 36.4 & 53.4\\ 
        \hline
        Zero Rate & 26.2 & 0.0 & - Baseline\\ 
        \bottomrule
    \end{tabularx}
    \caption{\pabi informativeness measures (x100) of \weakcq and other target tasks w.r.t \CQ and $\delta$-NLI. \blreded{\textbf{Bold}} values represent the maximum achievable \pabi Score by considering train subset as an \emph{indirect} signal for test subset of respective data. The highest \pabi score, excluding the max achievable, is indicated in \blpinked{\textit{italic}}.
    }
    \label{tab:pabi-eval}
    \vspace{-1.5em}
\end{table}

\begin{figure}
    \centering
    \includegraphics[width=0.9\linewidth]{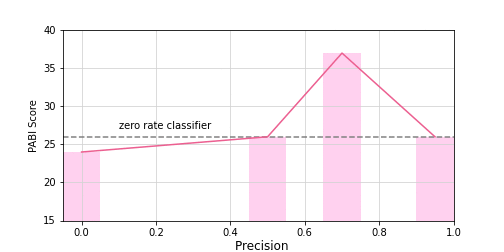}
    \caption{\pabi informativeness measures of \weakcq with different \recall thresholds on \CQ.}
    \label{fig:recal-eval}
    \vspace{-1.5em}
\end{figure}

\paragraph{Effect of \Recall on Informativeness:}
\Cref{fig:recal-eval} presents the \pabi informativeness estimation on weak supervision data under different threshold levels of \recall values, and compare them with the \QT{zero rate} classifier (always predicting majority class).
As illustrated, the informativeness show a significant drop in lower \recall showcasing the importance of using high \recall templates in our weak-supervision task.
For higher thresholds ($0.95$) the data will mostly consist of \emph{allow} patterns, the model drops to near zero rate informativeness baseline again.
This susceptibility on pattern \recall can be mitigated with having more fine-grained patterns on larger corpora. 
We leave further analysis on \recall of patterns to future work.

    \section{Analysis on Weak Supervision}
    \label{sec:datastats}
    In this section, we shift focus from external evaluation of \weakcq on target tasks to analyze distinct technical component of \weakcq. 
Here, through an ablation study, we try to answer four main questions to get more insight on the weak supervision provided by those components. 
First (Q1), how each labeling function (LF; \Cref{subsec:labeling}) is contributing to the extracted preconditions?
Second (Q2), what is the quality of the weak supervision data obtained from different ways of data acquisition?
Third (Q3), how does generative data augmentation (\Cref{subsec:augmentation}) contribute to \weakcq?
And finally (Q4), how much does the precondition-aware masking (\Cref{subsec:modified-mlm}) affect the overall performance of \weakcq?

\paragraph{(Q1) LF Analysis:}
To address the first question, we use statistics of the 6 top performing LFs (see \Cref{sec:details-on-LFs} for detailed results).
These 6 top performing LFs generate more than 80\% of data (Coverage) with the highest one generating 59\% of data and lowest one generating 1\%. 
Our results show that, in 0.14\% of instances we have conflict among competing LFs with different labels and in 0.12\% we have overlap among LFs with similar labels, which showcases the level of independence each LF has on individual samples.\footnote{Convectional inner-annotator agreement (IAA) methods hence are not applicable.} 

\paragraph{(Q2) Quality Control:}
To assess the quality of collected data, we used an expert annotator. 
The expert annotator is given a subset of the collected preconditions (preconditions-statement-label triplet) and asked to assign a binary label based on whether each the precondition is valid to its statement w.r.t the associated label.
We then report the average quality score as a proxy for \textit{precision} of data. 
We sampled 100 preconditions-statement-label triplets from three checkpoint in the pipeline: 1) extracted through linguistic patterns discussed in \Cref{subsec:labeling}, 2) outcome of the generative augmentations discussed in \Cref{subsec:augmentation}, and 3) final data used in \Cref{subsec:modified-mlm}.
Table~\Cref{tab:precisions} contains the average precision of the collected data, that shows the data has acceptable quality with minor variance in quality for different weak supervised steps in \weakcq.

\begin{table}[ht]
    \centering
    \small
    \begin{adjustbox}{max width=0.5\textwidth}
    \begin{tabular}{l|l}
        \midrule
        \textbf{Checkpoint Name} & \textbf{Precision. \%} \\ 
        \toprule
        Linguistic Patterns from~\Cref{subsec:labeling} & 78 \\
        Generative Augmentation from~\Cref{subsec:augmentation} & 76 \\
        Final Data used in \Cref{subsec:modified-mlm} & 76 \\
        \bottomrule
    \end{tabular}
    \end{adjustbox}
    
    \caption{Precision of the sampled preconditions-statement-label triplets from three checkpoints in pipeline.} 
    \label{tab:precisions}
    \vspace{-1.5em}
\end{table}

\paragraph{(Q3) Effectiveness of Generative Augmentation:}
The main effect of generative data augmentation (\Cref{subsec:augmentation}) is, among others, to acquire \weakcq additional training samples labeled as \emph{prevent} from pretrained LMs. 
When considering \CQ as target task, the \weakcq that does not use this technique (no-augment-\weakcq) sees a $4.14\%$ absolute drop in \fmac score. 
Upon further analysis of the two configurations, we observed that the no-augment-\weakcq leans more toward the zero rate classifier (only predicting \emph{allow} as the majority class) in comparison to the \weakcq.

\paragraph{(Q4) Effectiveness of Biased Masking:}
We focus on \CQ as the target task and compare the results of \weakcq with an alternative setup with no biased masking.
In the alternative setup, we only use the weak-supervision data obtained through \weakcq to fine-tune the model and compare the results.
Our results show that the \fmac score for zero-shot transfer learning setup has a $1.09\%$ absolute drop in \fmac score, without the biased masking process.

    \section{Related Work}
    \label{sec:related-work}
    \paragraph{Reasoning with Preconditions}
Collecting preconditions of \CSn and reasoning with them has been studied in multiple works.
\citet{rudinger2020thinking} uses the notion of \QT{defeasible inference}~\cite{pollock1987defeasible,levesque1990all} in term of how an \emph{update} sentence \textit{weakens} or \textit{strengthens} a \CSn hypothesis-premise pair.
For example, given the premise \QT{Two men and a dog are standing among rolling green hills.}, the \emph{update} \QT{The men are studying a tour map} weakens the hypothesis that \QT{they are farmers}, whereas \QT{The dog is a sheep dog} strengthens it.
Similarly, \CQ~\cite{qasemi2021corequisite} uses the notion of \QT{causal complex} from \citet{hobbs2005toward}, and defines preconditions as eventualities that either \textit{allow} or \textit{prevent} (allow negation~\cite{fikes1971strips} of) a \CSn \fact to happen.
For example, for the knowledge \QT{the glass is shattered} prevents the statement \QT{A glass is used for drinking water}, whereas "there is gravity" allows it.
In \CQ, based on  \citet{shoham1990nonmonotonic} and \citet{hobbs2005toward}, authors distinguish between two type of preconditions, causal connections (\emph{hard}), and material implication (tends to cause; \emph{soft}).
Our definition covers these definitions and is consistent with both.

\citet{hwang2020comet}, \citet{sap2019atomic}, \citet{heindorf2020causenet}, and \citet{speer2016conceptnet}, provided representations for preconditions of \facts in term of relation types, e.g. \emph{xNeed} in ATOMIC2020~\cite{hwang2020comet}.
However, the focus in none of these works is on evaluating SOTA models on such data.
The closest study of preconditions to our work are \citet{rudinger2020thinking}, \citet{qasemi2021corequisite}, \citet{do2021rotten} and \citet{jiang2021m}.
In these works, direct human supervision (crowdsourcing) is used to gather preconditions of \CSadj knowledge. They all show the shortcomings of SOTA models on dealing with such knowledge.
Our work differs as we rely on combination of distant-supervision and targeted fine-tuning instead of direct supervision to achieve on-par performance.
Similarly, \citet{mostafazadeh2020glucose}, and \citet{kwon-etal-2020-modeling} also study the problem of reasoning with preconditions.
However they do not explore \emph{preventing} preconditions.

\paragraph{Weak Supervision}
In weak-supervision, the objective is similar to supervised learning.
However instead of using human/expert resource to directly annotate unlabeled data, one can use the experts to design
user-defined patterns to infer \QT{noisy} or \QT{imperfect} labels~\cite{rekatsinas2017holoclean,zhang2017deepdive,dehghani2017neural,singh2022viphy}, e.g.\ using heuristic rules.
In addition, other methods such as re-purposing of external knowledge~\cite{alfonseca2012pattern,bunescu2007learning,mintz2009distant} or other types of domain knowledge~\cite{stewart2017label} also lie in the same category.
Weak supervision has been used extensively in NLU\@.
For instance, \citet{zhou2020temporal} utilize weak-supervision to extract temporal commonsense data from raw text, \citet{brahman2020learning} use it to generate reasoning rationale, \citet{dehghani2017neural} use it for improved neural ranking models, and \citet{hedderich2020transfer} use it to improve translation in African languages.
Similar to our work, ASER~\cite{zhang2020aser} and ASCENT~\cite{nguyen2020advanced} use weak supervision to extract relations from unstructured text.
However, do not explore preconditions and cannot express \emph{preventing} preconditions.
As they do focus on reasoning evaluation, the extent in which their contextual edges express \emph{allowing} preconditions is unclear.



\paragraph{Generative Data Augmentation}
Language models can be viewed as knowledge bases that implicitly store vast knowledge on the world.
Hence querying them as a source of weak-supervision is a viable approach.
Similar to our work, \citet{wang-etal-2021-table-based} use LM-based augmentation for saliency of data in tables, \citet{meng-etal-2021-distantly} use it as a source of weak-supervision in named entity recognition, and \citet{dai2021ultra} use masked LMs for weak supervision in entity typing.

%

    \section{Conclusion}
    \label{sec:conclusion}
    In this work we presented \weakcq, as an improved method for preconditioned \CSadj reasoning which involves two techniques of weak supervision.
To maximize the effect of the weak supervision data, we modified the masked language modeling loss function using biased masking method to put more emphasis on conjunctions as closest proxy to preconditions.
Through empirical and theoretical analysis of \weakcq, we show it significantly improves the results across the benchmarks on reasoning with the preconditions of commonsense knowledge.
In addition, we show the results are robust in different \recall values using the \pabi informativeness measure and
extensive ablation study.

Future work can consider improving the robustness of preconditioned inference models using methods such as virtual adversarial training \cite{miyato2018virtual,li2020tavat}.
With advent of visual-language models such as \citet{li2019visualbert}, preconditioned inference should also expand beyond language and include different modalities (such as image or audio).
To integrate in down-steam tasks, one direction is to include such models in aiding inference in the neuro-symbolic reasoners \cite{lin2019kagnet,verga2020facts}.

    \section*{Ethical Consideration}
    We started from openly available data that is both crowdsource-contributed and neutralized, however they still may reflect human biases. 
For example in case of \CQ~\cite{qasemi2021corequisite} they use ConceptNet as source of commonsense statements which multiple studies have shown its bias and ethical issues, e.g. ~\cite{mehrabi2021lawyers}.

During design of labeling functions we did not collect any sensitive information and the corpora we used were both publicly available, however they may contain various types of bias.
The labeling functions in \weakcq are only limited to English language patterns, which may inject additional cultural bias to the data.
However, our expert annotators did not notice any offensive language in data or the extracted preconditions.
Given the urgency of addressing climate change we have reported the detailed model sizes and runtime associated with all the experiments in Appendix~\ref{sec:model-sizes-and-run-times}.
    
    \section*{Limitations}
    The main limitation of this work are related to the choice of raw text corpora and the model for main results. 
From the raw text corpora perspective, we relied on Open Mind Common Sense (OMCS)~\cite{singh2002open} and ASCENT~\cite{nguyen2021advanced} as two rich resource of commonsense knowledge.
Future iterations of this work should include more fine-grained labeling functions to be applied to other large scale corpora that results in more diverse set of extracted preconditions.

The purpose of the experiments in this work is to show the effectiveness of \weakcq in preconditioned inference without introducing any expensive (manually labeled) supervision. 
We chose RoBERTa-Large-MNLI~\cite{liu2019roberta} as a representative and strong model that has been widely applied to NLI tasks, including all those evaluated in this work. 
However, there are more models, e.g. unified-QA-11B for \CQ or DeBERTa for $\delta$-NLI, that can be considered for each one of the target tasks. 
Of course achieving the SOTA with these much larger models requires a lot of computational resources, which is beyond the scope and bandwidth of this study. 
But, given more resources we would easily extend analysis to other models as well.

    \section*{Acknowledgement}
    We like to thank Hangfeng He, for his insightful comments to using the \pabi~\cite{forseeing} measure in our paper. We also want to thank our anonymous reviewers whose comments/suggestions helped improve and clarify this paper.
    
    This work is supported in part by the DARPA MCS program under Contract No. N660011924033 with the United States Office Of Naval Research, 
the National Science Foundation of United States
Grant IIS 2105329, and a Cisco Research Award.

    
    \bibliographystyle{acl_natbib}
    \bibliography{acl2021}

    \clearpage
	\setcounter{page}{1}
    \appendix
    
    \section{Details on \weakcq Method}
    \label{sec:details-on-weakcq-method}
    In this section, we discuss some of the extra details related to \weakcq and its implementation.

\subsection{Linguistic Patterns for \weakcq}
\label{subsec:appendx-patterns}
We use a set of conjunctions to extract sentences that follow the action-precondition sentence structure.
Initially, we started with two simple conjunctions-\textit{if} and \textit{unless}, for extracting assertions containing \emph{Allowing} and \emph{Preventing} preconditions, respectively.
To further include similar sentences, we expanded our vocabulary by considering the synonyms of our initial conjunctions.
Adding the synonyms of \textit{unless} we got the following set of new conjunctions for \emph{Preventing} preconditions-\{\textit{but, except, except for, if not, lest, unless}\}, similarly we expanded the conjunctions for Enabling preconditions using the synonyms of \textit{if}-\{\textit{contingent upon, in case, in the case that, in the event, on condition, on the assumption, supposing}\}.
Moreover, on manual inspection of the OMCS and ASCENT datasets, we found the following conjunctions that follow the Enabling precondition sentence pattern-\{\textit{makes possible, statement is true, to understand event}\}.
~\Cref{tab:patterns}, summarizes the final patterns used in \weakcq, coupled with their \recall value and their
associated conjunction.
\begin{table*}[ht]
    \centering
    \small
    \begin{adjustbox}{max width=\textwidth}
        \begin{tabular}{l|l|l}
            \midrule
            \textbf{Conjunctions} & \textbf{\Recall} & \textbf{Pattern}
            \\ \toprule
                but&  0.17& \{action\} but \{negative\_precondition\}\\
                contingent upon&  0.6& \{action\} contingent upon \{precondition\}\\
                except&  0.7& \{action\} except \{precondition\}\\
                except for&  0.57& \{action\} except for \{precondition\}\\
                if&  0.52& \{action\} if \{precondition\}\\
                if not&  0.97& \{action\} if not \{precondition\}\\
                in case&  0.75& \{action\} in case \{precondition\}\\
                in the case that&  0.30& \{action\} in the case that \{precondition\}\\
                in the event&  0.3& \{action\} in the event \{precondition\}\\
                lest&  0.06& \{action\} lest \{precondition\}\\
                makes possible&  0.81& \{precondition\} makes \{action\} possible.\\
                on condition&  0.6& \{action\} on condition \{precondition\}\\
                on the assumption&  0.44& \{action\} on the assumption \{precondition\}\\
                statement is true&  1.0& The statement "\{event\}" is true because \{precondition\}.\\
                supposing&  0.07& \{action\} supposing \{precondition\}\\
                to understand event&  0.87& To understand the event "\{event\}", it is important to know that \{precondition\}.\\
                unless&  1.0& \{action\} unless \{precondition\}\\
                with the proviso&  -& \{action\} with the proviso \{precondition\}\\
                on these terms&  -& \{action\} on these terms \{precondition\}\\
                only if&  -& \{action\} only if \{precondition\}\\
                make possible&  -& \{precondition\} makes \{action\} possible.\\
                without&  -& \{action\} without \{precondition\}\\
                excepting that&  -& \{action\} excepting that \{precondition\}
            \\\bottomrule    
        \end{tabular}
    \end{adjustbox}
    \caption{Linguistic patterns in \weakcq and their recall value. For patterns with not enough match in the corpora have empty recall values.} 
    \label{tab:patterns}
    \vspace{-1.0em}
\end{table*}

\subsection{Details of Snorkel Setup}
\label{subsec:appendx-snorkel}
Beyond a simple API to handle implementing patterns and applying them to the data, Snorkel's main purpose is to model and integrate noisy signals contributed by the labeling functions modeled as noisy, independent voters, which commit mistakes uncorrelated with other LFs.

To improve the predictive performance of the model, Snorkel additionally models statistical relationships between LFs. For instance, the model takes into account similar heuristics expressed by two LFs to avoid "double counting" of voters.
Snorkel, further, models the generative learner as a factor graph.
A labeling matrix $\Lambda$ is constructed by applying the LFs to unlabeled data points.
Here, $\Lambda_{i,j}$ indicates the label assigned by the $j^{th}$ LF for the $i^{th}$ data point.
Using this information, the generative model is fed signals via three factor types, representing the labeling propensity, accuracy, and pairwise correlations of LFs.

$\phi^{Lab}_{i,j}(\Lambda)=\mathds{1}\{\Lambda_{i,j} \neq \emptyset\}$

$\phi^{Acc}_{i,j}(\Lambda)=\mathds{1}\{\Lambda_{i,j} = y_{i}\}$

$\phi^{Corr}_{i,j,k}(\Lambda)=\mathds{1}\{\Lambda_{i,j} = \Lambda_{i,k} \}$

The above three factors are concatenated along with the potential correlations existing between the LFs and are further fed to a generative model which minimizes the negative log marginal likelihood given the observed label matrix $\Lambda$.

\subsection{Modified Masked Language Modeling}
\label{subsec:appendx-ModMLM}
\Cref{tab:conjuctions} summarizes the list of \emph{Allowing} and \emph{Preventing} conjunctions which the modified language modeling loss function is acting upon.
\begin{table*}[ht]
    \centering
    \small
        \begin{tabular}{l|p{0.85\textwidth}}
            \midrule
            \textbf{Type} 
                & \textbf{Conjunctions} 
            \\ \toprule
            \greened{Allowing}
                & only if, subject to, in case, contingent upon, given, if, in the case that, in case, in the case that, in the event, on condition, on the assumption, only if, so, hence, consequently, on these terms, subject to, supposing, with the proviso, so, thus, accordingly, therefore, as a result, because of that, as a consequence, as a result
            \\ 
            \reded{Preventing}
                & but, except, except for, excepting that, if not, lest, saving, without, unless
            \\\bottomrule    
        \end{tabular}
    \caption{List of conjunctions used in modified masked loss function in section~\ref{subsec:modified-mlm}} \label{tab:conjuctions}
    \vspace{-1.0em}
\end{table*}

\begin{table}[ht]
\centering
\small
\begin{tabularx}{0.5\textwidth}{l|X}
    \midrule
    \textbf{Conjunction} & \textbf{Pattern} \\ 
    \toprule
    \greened{to understand event} & To understand the event ``\{event\}", it is important to know that \{precondition\}. \\ \hline
    \greened{in case}             & \{action\} in case \{precondition\} \\ 
    \hline
    \greened{statement is true}   & The statement ``\{event\}" is true because \{precondition\}. \\ 
    \hline
    \reded{except}                & \{action\} except \{precondition\}                                                   \\ 
    \hline
    \reded{unless}               & \{action\} unless \{precondition\}                                                   \\ \hline
    \reded{if not}               & \{action\} if not \{precondition\}                                                   \\ \bottomrule
\end{tabularx}
\caption{Filtered Labeling Functions Patterns and their associated polarity.} \label{tab:lf_patterns}
\vspace{-1.5em}
\end{table}

\subsection{Interrogative Words}
\label{subsec:interrogative-words}
On manual inspection of the dataset, we observed some sentences that were not relevant to the common sense reasoning task. Many of such instances were interrogative statements. We filter out such cases based on the presence of interrogative words in the beginning of a sentence. These interrogative words are listed below.

Interrogative words: ["Who", "What", "When", "Where", "Why", "How", "Is", "Can", "Does", "Do"]

    \section{Details on Target Data Experiments}
    \label{sec:details-on-target-data-experiments}
    \label{appendix:target-datasets}
For converting \citet{rudinger2020thinking}, similar to \citet{qasemi2021corequisite}, we concatenate the \QT{Hypothesis} and \QT{Premise} and consider then as NLI's hypothesis.
We then use the \QT{Update} sentence as NLI's premise.
The labels are directly translated based on \emph{Update} sentences's label, \emph{weakener} to \emph{prevent} and the \emph{strengthener} to \emph{allow}.

To convert the ATOMIC2020~\cite{hwang2020comet}, similar to \citet{qasemi2021corequisite}, we focused on three relations \emph{HinderedBy}, \emph{Causes}, and \emph{xNeed}.
From these relations, edges with \emph{HinderedBy} are converted as \emph{prevent} and the rest are converted as \emph{allow}.

Winoventi~\cite{do2021rotten}, proposes Winograd-style {ENTAILMENT} schemas focusing on negation in \CSn.
To convert it to NLI style, we first separate the two sentences in the \textit{masked\_prompt} of each instance to form \emph{hypothesis} and \emph{premise}.
We get two versions of \emph{premise} by replacing the MASK token in \emph{premise} with their \emph{target} or \emph{incorrect} tokens.
For the labels the version with \emph{target} token is considered as \emph{allow} and the version with \emph{incorrect} token as \emph{prevent}.

ANION~\cite{jiang2021m}, focuses on {CONTRADICTION} in general.
We focus on their commonsense d{CONTRADICTION} subset as it is clean of lexical hints.
Then we convert their crowdsourced \emph{original head} or \emph{{CONTRADICTION} head} as hypothesis, and the lexicalized predicate and tail as the premise (e.g. \emph{xIntent} to \emph{PersonX intends to}).
Finally the label depends on head is \emph{allow} for \emph{original head} and \emph{prevent} for \emph{{CONTRADICTION} head}.
We also replace \QT{PersonX} and \QT{PersonY} with random human names (e.g. \QT{ALice}, \QT{Bob}).

Finally, for the \CQ~\cite{qasemi2021corequisite}, we used their proposed P-NLI task as a NLI-style task derived from their preconditions dataset.
We converted their \emph{Disabling} and \emph{Enabling} labels to \emph{prevent} and \emph{allow} respectively.

\Cref{tab:target-to-NLI} summarizes the conversion process through examples from the original data and the NLI task derived from each.

\begin{table*}[ht]
    \centering
    \small
    \begin{adjustbox}{max width=\textwidth}
        \begin{tabular}{l|ll|ll}
            \midrule
            \textbf{Name} 
                & \textbf{Original Data} &
                & \textbf{Derived NLI} &
            \\ \toprule
            \multirow{3}{*}{\shortstack[l]{Winoventi \\ \cite{do2021rotten}}} 
            & 
            \shortstack[l]{\textbf{masked\_prompt}:\\\textcolor{white}{a}} & \shortstack[l]{Margaret smelled her bottle of maple syrup \\ and it was sweet. The syrup is \{MASK\}.}
            & \shortstack[l]{\textbf{Hypothesis}:\\\textcolor{white}{a}} & \shortstack[l]{Margaret smelled her bottle of maple syrup \\ and it was sweet.}
            \\ & \textbf{target}: & edible 
            & \textbf{Premise}: & The syrup is edible/malodorous
            \\ & \textbf{incorrect}: & malodorous 
            & \textbf{Label}: & \greened{ENTAILMENT}/\reded{CONTRADICTION}
            \\ \hline
            \multirow{4}{*}{\shortstack[l]{ANION \\ \cite{jiang2021m}}} 
            & 
            \textbf{Orig\_Head}: & PersonX expresses PersonX's delight.
            & \textbf{Hypothesis}: & Alice expresses Alice's delight/anger. 
            \\ & \textbf{Relation}: & xEffect 
            & \textbf{Premise}: & feel happy.
            \\ & \textbf{Tail}: & Alice feel happy 
            & \textbf{Label}: & \greened{ENTAILMENT}/\reded{CONTRADICTION}
            \\ & \textbf{Neg\_Head}: & PersonX expresses PersonX's anger. 
            &
            \\ \hline
            \multirow{3}{*}{\shortstack[l]{ATOMIC2020 \\ \cite{hwang2020comet}}} 
            & 
            \textbf{Head}: & PersonX takes a long walk.
            & \textbf{Hypothesis}: & PersonX takes a long walk. 
            \\ & \textbf{Relation}: & HinderedBy 
            & \textbf{Premise}: & It is 10 degrees outside..
            \\ & \textbf{Tail}: & It is 10 degrees outside. 
            & \textbf{Label}: & \reded{CONTRADICTION}
            \\ \hline
            \multirow{4}{*}{\shortstack[l]{$\delta$-NLI \\ \cite{rudinger2020thinking}}}
            & 
            \textbf{Hypothesis}: & PersonX takes a long walk.
            & \textbf{Hypothesis}: & PersonX takes a long walk. 
            \\ & \textbf{Premise}: & HinderedBy 
            & \textbf{Premise}: & It is 10 degrees outside..
            \\ & \textbf{Update}: & It is 10 degrees outside. 
            & \textbf{Label}: & \reded{CONTRADICTION}
            \\ & \textbf{Label}: & Weakener &
            \\ \hline
            
            \multirow{3}{*}{\shortstack[l]{\CQ \\ \cite{qasemi2021corequisite}}} 
            & 
            \textbf{Statement}: & A net is used for catching fish.
            & \textbf{Hypothesis}: & A net is used for catching fish. 
            \\ & \textbf{Precondition}: & You are in a desert. 
            & \textbf{Premise}: & You are in a desert.
            \\ & \textbf{Label}: & Disabling 
            & \textbf{Label}: & \reded{CONTRADICTION}
            \\ \bottomrule
        \end{tabular}
    \end{adjustbox}
    \caption{Examples from target tasks in NLI format} 
    \label{tab:target-to-NLI}
    \vspace{-1.5em}
\end{table*}

To run all the experiments, we fine-tune the models on tuning data for maximum of 5 epochs with option for early stopping available upon 5 evaluation cycles with less than $1e-3$ change on validation data. For optimizer, we use AdamW~\cite{adamw} with learning rate of 3e-6 and default hyperparamter for the rest. 
    
    \section{Curriculum vs. Multitask Learning}
    \label{subsec:curic-vs-multitask}
    For results of \Cref{subsec:target-tasks}, we considered the target task and \weakcq as separate datasets, and fine-tuned model sequentially on them (curriculum learning;\citealp{pentina2015curriculum}).
We chose \emph{curriculum} learning setup due to its simplicity in implementation, ease of fine-tuning process monitoring and hyperparameter setup. 
It would also allow us to monitor each task separately that increases interpretability of results.

However, in an alternative fine-tuning setup, one can merge the two datasets into one and fine-tune the model on the aggregate dataset (multi-task learning;\citealp{caruana1997multitask}).
Here, we investigate such alternative and its effect on the results of \Cref{subsec:target-tasks}.

\paragraph{Setup} We use the same setup as \Cref{subsec:target-tasks} for fine-tuning the model on \emph{Orig.+\weakcq}. 
Here instead of first creating \weakcq and then fine-tuning it on the target task, we merge the weak-supervision data of \weakcq with the training subset of the target task and then do fine-tuning on the aggregate dataset.
To manage length of this section, we only consider \CQ, $\delta$-NLI and Winoventi as the target dataset.

\begin{table}[ht]
    \small
    \centering
    \begin{tabular}{l|ll}
     \midrule
     \textbf{Target Data}
         & \textbf{Orig+\weakcq (Multi-Task)}
         & \textbf{Diff.}
     \\ 
     \toprule
     $\delta$-NLI & 72.1 & -11.00 \\
     \CQ           & 77.3 & +6.8   \\
     Winoventi    & 51.7 & +0.7   \\
     \bottomrule
    \end{tabular}
    \caption{\fmac(x100) results of \weakcq on the target datasets using \emph{multi-task} fine-tuning strategy and its difference with \emph{curriculum} strategy.}
    \label{tab:curic-results}
    \vspace{-1.5em}
\end{table}

\paragraph{Discussion}
\Cref{tab:curic-results} summarizes the results for \emph{multi-task} learning setup and its difference w.r.t to the results of the \emph{curriculum} learning setup in \Cref{tab:target-eval}. 
Using \emph{multi-task} learning does not show the consistent result across tasks. 
We see significant performance loss on $\delta$-NLI on one hand and major performance improvements on \CQ on the other.
The Winoventi, however appears to not change as much in the new setup.
We leave further analysis of \emph{curriculum learning} to future work.
    
    \section{Model Sizes and Run-times}
    \label{sec:model-sizes-and-run-times}
    All the experiments are conducted on a commodity workstation with an Intel Xeon Gold 5217 CPU and an NVIDIA RTX 8000 GPU.
For all the fine-tuning results in \Cref{tab:target-eval}, \Cref{tab:pabi-eval} we used \QT{RoBERTa-Large-MNLI} with 356M tuneable parameters.
To fine-tune the model in each experiment, we use Ray~\cite{liaw2018tune} to handle hyperparameter tuning with 20 samples each. 
The hyperparameters that are being tuned fall into two main categories: 1) model hyperparameters such as \QT{sequence length}, \QT{batch size}, etc. and 2) data hyperparameters such as \QT{precision threshold}, \QT{data size}, etc..
The mean run-time for each sample on target datasets is 1hr 55mins.
For the augmentation in \weakcq dataset, we used \QT{BERT} language model with $234M$ tuneable parameters.
The mean run-time on the weak supervision data is 49hr that includes all three steps of data preprocessing, linguistic pattern matching, and generative data augmentation.

    \section{Details on \pabi Measurement}
    \label{sec:details-on-pabi}

\pabi provides an Informativeness measure that quantifies the reduction in uncertainty provided by incidental supervision signals. We use the \pabi measure to study the impact of transductive cross-domain signals obtained from our weak-supervision approach.

Following \cite{forseeing}, in order to calculate \pabi $\hat{S}(\pi_{0},\tilde{\pi}_{0})$, we first find out $\eta$, the difference between a perfect system and a gold system in the target domain $\mathcal{D}$ that uses a label set $\mathcal{L}$ for a task, using Eq.\ref{eq:pabi1}. 
\vspace{-0.5em}
\begin{equation}
\label{eq:pabi1}
\begin{split}
    \eta&=\mathbb{E}_{x\sim P_{\mathcal{D}(x)}}1(c(x)\neq \tilde c(x)) \\
    &=\frac{(\lvert\mathcal{L}\rvert-1)(\eta_{1}-\eta_{2})}{1-\lvert\mathcal{L}\rvert(1-\eta_{1})}
\end{split}
\end{equation}
\begin{table*}[ht]
    \centering
    \small
    \begin{adjustbox}{max width=\textwidth}
    \begin{tabular}{l|lllll|lll|lll}\midrule
        Indir. Task & $|L|$ & $\eta_1$ & $\eta_2^{ATMC}$ & $\eta_2^{\CQ}$& $\eta_2^{\delta-NLI}$ &$\eta^{ATMC}$  &$\eta^{\CQ}$ &$\eta^{\delta-NLI}$ &$\pabi^{ATMC}$ &$\pabi^{\CQ}$ &$\pabi^{\delta-NLI}$ \\\toprule
        
        \weakcq   &2 &0.04 &0.11 &0.21 &0.16 &0.076 &0.202 &0.129 &0.782 &0.523 &0.667 \\
    $\delta$-NLI  &2 &0.13 &0.22 &0.28 &0.16 &0.122 &0.203 &0.046 &0.683 &0.522 &0.855 \\
        \CQ       &2 &0.03 &0.10 &0.22 &0.33 &0.074 &0.202 &0.318 &0.786 &0.523 &0.313 \\
        ATOMIC    &2 &0.01 &0.57 &0.62 &0.60 &0.608 &0.622 &0.602 &0.184 &0.209 &0.174 \\
        ANION     &2 &0.16 &0.57 &0.36 &0.44 &0.571 &0.302 &0.418 &0.122 &0.341 &0.139 \\
        Winoventi &2 &0.19 &0.10 &0.37 &0.31 &0.139 &0.289 &0.196 &0.647 &0.364 &0.534 \\
        \bottomrule
    \end{tabular}
    \end{adjustbox}
    \caption{Details of \pabi metric computations in \Cref{subsec:info-eval} according to \Cref{eq:pabi1}}
    \label{tab:pabi-details}
    
\end{table*}
Here, $P_{\mathcal{D}(x)}$ indicates the marginal distribution of $x$ under $\mathcal{D}$, $c(x)$ refers to gold system on gold signals, $\tilde c(x)$ is a perfect system on incidental signals, $\eta_{1}$ refers to the difference between the silver system and the perfect system in the source domain, $\acute{\eta_{1}}$ indicates difference between the silver system and the perfect system in the target domain, and $\eta_{2}$ is the difference between the silver system and the gold system in the target domain.

Using Eq.\ref{eq:pabi1}, the informative measure supplied by the transductive signals $\hat{S}(\pi_{0},\tilde{\pi}_{0})$ can be calculated as follows:
$$\sqrt{1-\frac{\eta\ln(\lvert\mathcal{L}\rvert-1)-\eta\ln\eta-(1-\eta)\ln(1-\eta))}{\ln\lvert\mathcal{L}\rvert}}$$

\Cref{tab:pabi-details} contains the details associated computation of \pabi score as reported in \Cref{subsec:info-eval}.

    \section{Details on LFs in \weakcq}
    \label{sec:details-on-LFs}
    \Cref{tab:combined_lfa} shows Coverage (fraction of instances assigned the non-abstain label by the labeling function), Overlaps (fraction of instances with at least two non-abstain labels), and Conflicts (fraction of instances with conflicting and non-abstain labels) on top performing LFs in \weakcq.
\begin{table}[h]
\centering
\small
\begin{adjustbox}{max width=0.5\textwidth}
\begin{tabular}{l|l|l|l}
    \midrule
    \textbf{LF name} & \textbf{Cov. \%} & \textbf{Over. \%} & \textbf{Conf. \%} \\ 
    \midrule
    \greened{to understand} &59.03 &0.03 &0.03 \\
    \greened{statement is}&10.58 &0.03 &0.03 \\
    \reded{except} &4.84 &0.02 &0.01 \\
    \reded{unless} &4.79 &0.04 &0.04 \\
    \greened{in case} &1.46 &0.01 &0.00 \\
    \reded{if not} &1.00 &0.01 &0.01 \\ \hline
    Overall &81.69 &0.14 &0.12 \\
    \bottomrule
\end{tabular}
\end{adjustbox}
\caption{Coverage (fraction of
raw corpus instances assigned the non-abstain label by the labeling function), Overlaps (fraction of raw corpus instances with at least two non-abstain labels), and Conflicts (fraction of the raw corpus instances with conflicting (non-abstain) labels) on top performing LFs.  \greened{Green} and \reded{red} color respectively represent LFs that assign \emph{allow} and \emph{prevent} labels.
} 
\label{tab:combined_lfa}
\vspace{-1.5em}
\end{table}

    \section{Details on Preconditioned Inference in the Literature}
    \label{sec:other_prec_inference}
    As mentioned in \cref{sec:background}, existing literature does not have a consistent (unified) definitions from to aspects: 1) the definition of the preconditions, and 2) the definition of preconditioned inference. 

First, existing literature define preconditions of \CSn \facts in different degrees of impact on the \fact.
For example, \citet{qasemi2021corequisite} follows the notion of \QT{causal complex} from \citet{hobbs2005toward}, where for a \CSn \fact~$s$ preconditions of the \fact~$P_f(s)$ are defined as collection of eventualities (events or states) that results in $s$ to happen.
According to \citet{qasemi2021corequisite}, such eventualities can either \textit{enable}~($p_{f}^+ \in P_f$) or \textit{disable}~($p_{f}^- \in P_f$) the \fact to happen. Also, \citet{qasemi2021corequisite} uses \citet{fikes1971strips} to define \textit{disable} as \textit{enabl}ing the negation of the \fact.
On other hand, \citet{rudinger2020thinking} defines \textit{strengthener} as updates that a human would find them to increase likelihood of a hypothesis, and the \textit{weakener} as the one that humans would find them to decrease it. 
Here, the focus on human's opinion is stemmed from definition of \CSn.
In this work, given the focus on noisy labels derived from weak-supervision, we adopted the more relaxed definition from \citet{rudinger2020thinking} for preconditions of \CSn \facts.

Second, there is also inconsistencies in the definition of reasoning with the preconditions or preconditioned inference. 
\citet{rudinger2020thinking} has a strict structure. It defines the task w.r.t to effect of precondition on the relation of two sentences: hypothesis and premise; where a model has to find the type of the precondition based on whether it \textit{strengthens} or \textit{weakens} the relation between the two sentences.
Differently, \citet{qasemi2021corequisite} has a relaxed definition in which the model is to decide if the precondition either \text{enables} or \textit{disables} the \fact. Here the \fact can have any format. 
\citet{do2021rotten}, \citet{hwang2020comet}, and \citet{jiang2021m}, on the other hand, define only a generative task to evaluate the models.
In this work, again we adopted the more relaxed definition from \citet{qasemi2021corequisite} that imposes less constraint on weak-supervised data.

\end{document}